\documentclass[graybox, envcountchap]{svmult}

\usepackage{mathptmx}        
\usepackage{helvet}          
\usepackage{courier}         

\usepackage{makeidx}         
\usepackage{graphicx}        
\usepackage{multicol}        
\usepackage[bottom]{footmisc}

\makeindex             

\begin{document}

\frontmatter

%
%
%

\begin{dedication}
Use the template \emph{dedic.tex} together with the Springer document class SVMono for monograph-type books or SVMult for contributed volumes to style a quotation or a dedication\index{dedication} at the very beginning of your book in the Springer layout
\end{dedication}

%
%

\foreword

Use the template \textit{foreword.tex} together with the Springer document class SVMono (monograph-type books) or SVMult (edited books) to style your foreword\index{foreword} in the Springer layout. 

The foreword covers introductory remarks preceding the text of a book that are written by a \textit{person other than the author or editor} of the book. If applicable, the foreword precedes the preface which is written by the author or editor of the book.

\vspace{\baselineskip}
\begin{flushright}\noindent
Place, month year\hfill {\it Firstname  Surname}\\
\end{flushright}

%
%

\preface

Use the template \emph{preface.tex} together with the Springer document class SVMono (monograph-type books) or SVMult (edited books) to style your preface in the Springer layout.

A preface\index{preface} is a book's preliminary statement, usually written by the \textit{author or editor} of a work, which states its origin, scope, purpose, plan, and intended audience, and which sometimes includes afterthoughts and acknowledgments of assistance. 

When written by a person other than the author, it is called a foreword. The preface or foreword is distinct from the introduction, which deals with the subject of the work.

Customarily \textit{acknowledgments} are included as last part of the preface.

\vspace{\baselineskip}
\begin{flushright}\noindent
Place(s),\hfill {\it Firstname  Surname}\\
month year\hfill {\it Firstname  Surname}\\
\end{flushright}

%
%

\extrachap{Acknowledgements}

Use the template \emph{acknow.tex} together with the Springer document class SVMono (monograph-type books) or SVMult (edited books) if you prefer to set your acknowledgement section as a separate chapter instead of including it as last part of your preface.

\tableofcontents
%
%
%
\contributors

\begin{thecontriblist}
Firstname Surname
\at ABC Institute, 123 Prime Street, Daisy Town, NA 01234, USA, \email{smith@smith.edu}
\and
Firstname Surname
\at XYZ Institute, Technical University, Albert-Schweitzer-Str. 34, 1000 Berlin, Germany, \email{meier@tu.edu}
\end{thecontriblist}
%
%

\extrachap{Acronyms}

Use the template \emph{acronym.tex} together with the Springer document class SVMono (monograph-type books) or SVMult (edited books) to style your list(s) of abbreviations or symbols in the Springer layout.

Lists of abbreviations\index{acronyms, list of}, symbols\index{symbols, list of} and the like are easily formatted with the help of the Springer-enhanced \verb|description| environment.

\begin{description}[CABR]
\item[ABC]{Spelled-out abbreviation and definition}
\item[BABI]{Spelled-out abbreviation and definition}
\item[CABR]{Spelled-out abbreviation and definition}
\end{description}

\mainmatter
%
%
%

\begin{partbacktext}
\part{Part Title}
\noindent Use the template \emph{part.tex} together with the Springer document class SVMono (monograph-type books) or SVMult (edited books) to style your part title page and, if desired, a short introductory text (maximum one page) on its verso page in the Springer layout.

\end{partbacktext}

\title*{GraphWOZ: Dialogue Management with Conversational Knowledge Graphs}
\author{Nicholas Thomas Walker, Stefan Ultes, and Pierre Lison}
\institute{Nicholas Walker \at Norwegian Computing Center, Postboks 114 Blindern, \email{walker@nr.no}
\and Stefan Ultes \at University of Bamberg,  96045 Bamberg, Germany, \email{stefan.ultes@uni-bamberg.de}
\and Pierre Lison \at Norwegian Computing Center, Postboks 114 Blindern, \email{plison@nr.no}}
%
%
\maketitle

\abstract{We present a new approach to dialogue management using conversational knowledge graphs as core representation of the dialogue state. 
To this end, we introduce a new dataset, GraphWOZ, which comprises Wizard-of-Oz dialogues in which human participants interact with a robot acting as a receptionist. 
In contrast to most existing work on dialogue management, GraphWOZ relies on a dialogue state explicitly represented as a dynamic knowledge graph instead of a fixed set of slots. 
This graph is composed of a varying number of entities (such as individuals, places, events, utterances and mentions) and relations between them (such as persons being part of a group or attending an event). 
The graph is then regularly updated on the basis of new observations and system actions. 
GraphWOZ is released along with detailed manual annotations related to the user intents, system responses, and reference relations occurring in both user and system turns.
Based on GraphWOZ, we present experimental results for two dialogue management tasks, namely conversational entity linking and response ranking. For conversational entity linking, we show how to connect utterance mentions to their corresponding entity in the knowledge graph with a neural model relying on a combination of both string and graph-based features. 
Response ranking is then performed by summarizing the relevant content of the graph into a text, which is concatenated with the dialogue history and employed as input to score possible responses to a given dialogue state.}

\section{Introduction}
\label{sec:Introduction}
Task-oriented dialogue systems often include a dialogue management module in charge of tracking relevant information and selecting the next system response or action to perform \cite{henderson2015machine}. This module often revolves around a representation of the current dialogue state comprising domain-specific and predefined ``slots'' that are necessary to complete a task within a domain.

While most previous work on dialogue management renders the dialogue state with a fixed list of predefined slots, this paper follows an alternative approach based on a dialogue state expressed as a graph of conversational entities \cite{ultes2018addressing,10.1007/978-981-19-5538-9_15}. Graphs are increasingly employed in dialogue research, albeit typically as static graphs of background knowledge \cite{madotto2020learning, young2018augmenting,zhou2018commonsense} rather than to capture the entirety of the dialogue state. Such knowledge graphs are frequently used with open domain dialogue systems, but have also been integrated with neural dialogue models for task-oriented dialogue systems \cite{ghazvininejad2018knowledge}

Graph representations of dialogue state are particularly relevant for domains that must capture a rich conversational context, as in virtual assistants or human--robot interaction. Graphs makes it possible to directly express the connections between abstract entities and their lower-level groundings in the external environment, along with relations between those entities and the tasks to perform \cite{bonial2020dialogue}. 

Several datasets for dialogue-oriented semantic parsing have been released in recent years \cite{andreas2020task,bonial2020dialogue}. However, there is a lack of datasets that specifically target the problem of dialogue management with graph-structured dialogue states. This paper seeks to address this shortcoming, and presents GraphWOZ, a new Wizard-of-Oz dataset in which participants interact with a robot receptionist. The dialogue state underlying this domain includes a varying number of entities -- such as persons, events or locations -- connected with one another through relations (such as a person attending an event or being at a particular location). 

Along with this new dataset, we present novel approaches to perform two dialogue management tasks based on this graph-structured dialogue state:
\begin{enumerate}
\item A neural feed-forward model for conversational entity linking based on a combination of string and graph distance features ;
\item A simple but effective method for response selection, which operates by converting a relevant part of the graph into text and running a fine-tuned language model to rank candidate responses. 
\end{enumerate}

The rest of this paper is as follows. We first review in Section \ref{sec:related} related work on the use of graphs for dialogue management. Section \ref{sec:Collection} describes the creation and annotation of the GraphWOZ dataset. In Section \ref{sec:approach}, we present the representation of the dialogue state as a graph, and the use of this representation for conversational entity linking and response selection. Finally, Section \ref{sec:results} provides experimental results on those two tasks, and Section \ref{sec:conclusion} concludes.

\section{Related Work}
\label{sec:related}
Graphs have been used in numerous contexts in dialogue systems research. For instance, the Microsoft Concept Graph has been used to connect background information to entities in the dialogue, along with a state graph to manage overall dialogue flow \cite{pichl2018alquist}. Similarly, graphs have been used to graphs represent multi-turn utterances and intents \cite{yang2020reinforcement}, while other work has used dataflow graphs representing executable programs \cite{andreas2020task}. Key-Value Retrieval Networks made use of task-specific information structured as a Knowledge graph (the KVRET* dataset) \cite{eric2017key}. Dynamic knowledge graphs in particular have been used in in a Question Answering (QA) task to represent domain-slot pairs \cite{zhou2018commonsense}. The use of knowledge graphs has been prominent in QA, with uses including enhancement of pre-trained language models \cite{shen-etal-2020-exploiting}. The AgentGraph framework \cite{chen2019agentgraph} used graph neural networks to improve sample efficiency in learning a dialogue management policy, while ConvGraph \cite{gritta2021conversation} used a graph-based approach to dialogue data augmentation. Recently, graphs have also been used to represent both dialogue context and external knowledge in a property graph \cite{10.1007/978-981-19-5538-9_15} or to jointly encode dialogue context and a knowledge base for a graph neural network \cite{liu2021heterogeneous}. Graphs have also been used to encode relationships between visual and language data \cite{hong2020language} and mental states of dialogue agents \cite{qiu2022towards}.

Entity linking is a core task in many NLP applications \cite{mudgal2018deep}, and can be defined in the following way \cite{shen2014entity}:
\begin{quote}
    Given a knowledge base containing a set of entities \textit{E} and a text collection in which a set of named entity mentions \textit{M} are identified in advance, the goal of entity linking is to map each textual entity mention $m \in M$ to its corresponding entity $e \in E$ in the knowledge base
\end{quote}

Entity linking has been explored in various tasks in NLP, particularly in the rich literature on knowledge base construction \cite{ji2011knowledge}, conversational QA, and knowledge base QA \cite{lansurvey}. This often takes the form of retrieving information about entities from large scale knowledge bases \cite{shen-etal-2019-multi}. Other work in conversational QA has explored the use of Graph Attention Networks to model relationships between entities \cite{kacupaj-etal-2021-conversational}. Most existing work in entity linking has been focused on text, albeit few studies have also investigated entity linking on noisy ASR transcriptions \cite{benton2015entity}. 


Entity linking has also been explored in conversational settings, particularly for open domain dialogues \cite{cui-etal-2022-openel}.  The representation of these entities and their links to each other has taken various forms \cite{joko2021conversational}. So-called entity centric models focus on learning representations of dialogue entities \cite{aina2019entity}, without necessarily establishing links to a particular knowledge graph. An example is the Recurrent Entity, which Network \cite{henaff2017tracking} processes text data using a dynamic memory component to learn representations of dialogue. In such cases, it has been noted that there exists a close relationship between entity linking and coreference resolution \cite{chen2017robust}, and this relationship has been explored by modelling coreference resolution and entity linking as a joint task \cite{zheng-etal-2013-dynamic}. Similarly, Slugbot \cite{bowden2018slugnerds} explored Named-Entity Recogntion (NER) and entity linking with the Google Knowledge Graph. Other dialogue systems have explored the use of large knowledge graphs \cite{curry2018alana}. Recent work has also explored the problem of plural mentions \cite{zhou-choi-2018-exist}, and entity linking within multi-party dialogue, where it can be referred to as character identification \cite{chen2016character}. 

Previous work has additionally explored the use of document representations of semantic graphs \cite{li2022enhanced}, an approach which stands closely to our response ranking experiments. Hand crafted features as well as neural methods have been employed for response ranking \cite{Papaioannou2017AnEM, shalyminov-etal-2018-neural}. Task-oriented dialogue systems specifically have made use of ranking modules with document-grounded large pre-trained language models \cite{kim-etal-2021-document, tan2020learning}. The PolyResponse model relied on a non-generative response ranking component to present both language and visual data to the user \cite{henderson-etal-2019-polyresponse}. Other work has demonstrated improvements to dialogue ranking models by using synthesized negative training examples \cite{gupta-etal-2021-synthesizing}.

The GraphWOZ dataset was collected with a "Wizard-of-Oz" (WOZ) setup, which is a cost-effective method to collect dialogue data \cite{kelley1984iterative,wen2017network}. MultiWOZ \cite{budzianowski2018multiwoz} and Taskmaster \cite{byrne2019taskmaster}  are other datasets of task-oriented dialogues collected through WOZ experiments. 

\section{Data}
\label{sec:Collection}


\subsection{Dialogue collection}

GraphWOZ consists of interactions between a human participant that engages with a Pepper robot\footnote{\url{https://www.softbankrobotics.com/emea/en/pepper}} acting as a receptionist. The dialogue state contains both physical and abstract entities (such as persons, groups, events, locations and time references) that have relevance to the task, and which may change in the course of the dialogue. GraphWOZ is released along with annotations which reflect both user intents, system actions, and the operations performed on the entity graph at each turn. GraphWOZ is made publicly available on Github\footnote{\url{https://github.com/ntwalker/GraphWOZ}}.

At the onset of each dialogue, we generate a fictive organization composed of randomly generated entities. Each organization consists of between 40 and 60 fictive persons, and all individuals in the organization are assigned to a group and randomly generated office number along with a phone and email address. Subsequently, 30 to 50 events are generated with names combining randomly sampled jargon and a set of pre-defined meeting types (e.g. "web-readiness" + "status update"). Each event is randomly assigned to an organizer, a location and a start and end time. The event locations are randomly selected from a set of fictive conference rooms and the office of the person organizing the event. 

Each dialogue begins with a task shown to the user on the robot tablet. The tasks are instantiated using predefined templates, which contain typed placeholders for elements such as people, events, places, and times. These slots are filled in by randomly sampling the generated organization. Tasks consist of combinations of topics such as scheduling, moving, or cancelling meetings, or finding attributes of entities in the organization. After the user finishes a task, a new task is created and displayed. Participants were given flexibility to interpret the task as desired and mark it as completed or failed to the best of their knowledge. 
The participants were instructed to work with the robot to negotiate a reasonable solution matching the intention of the task to the best of their ability. When the task is marked by the participant as completed or aborted, a new task is generated. Each user utterance is transcribed using Google Speech and the top ASR transcription is shown to the wizard. The wizard then composes a free-text response to the user's utterance, which is uttered by the robot using text-to-speech synthesis as a response to the user.

An example of a user task is as follows:
\begin{quote}
    You are Richard Schroeder, and you want to arrange a meeting called methodologies discussion. Make a half hour long meeting in an available conference room, and invite Daniel Patton and at least one member of the Mathematics group.
\end{quote}

A total of 15 distinct speakers (university students or researchers in data science) participated in the data collection, and interacted with the robot receptionist for about an hour each. The interactions took place in English (most participants were non-native speakers of English but had a high command of the language). The participants were paid for their participation, and provided their consent to the inclusion of their dialogues in a publicly released dataset. 

\subsection{Data Encoding}
\label{sec:Data}


In total, GraphWOZ contains 119 dialogues from 15 distinct speakers. The dialogues are between 11 and 12 turns long on average. Figures \ref{fig:dialogue1} and \ref{fig:dialogue2} provide two concrete examples of dialogue in GraphWOZ.

\begin{figure*}[t]
    \FramedBox{3cm}{11.25cm}{
    \textbf{1} \textsc{Visitor:}\ \ Hello! My name is \textcolor{Green}{\dotuline{Wendy Parker}} and I am trying to find out who is organizing the \textcolor{Purple}{\dotuline{users workshop}}. \\
    \textbf{2} \textsc{Robot:}\ \ \ \ \ Hi \textcolor{Green}{\dotuline{Wendy}}! The \textcolor{Purple}{\dotuline{users workshop}} is organized by \textcolor{Green}{\dotuline{Mark Suarez}}. \\
    \textbf{3} \textsc{Visitor:}\ \ \ Could you give me the room number for \textcolor{Green}{\dotuline{his}} office? \\
    \textbf{4} \textsc{Robot:}\ \ \ \ \ Sure, \textcolor{Green}{\dotuline{his}} office is \textcolor{Orange}{\dotuline{room 270}}. \\
    \textbf{5} \textsc{Visitor:}\ \ Thank you, that's all. \\
    \textbf{6} \textsc{Robot:}\ \ \ \ \ Okay! Have a nice day! \\ }
    \caption{Short dialogue example with a fragment of the dialogue state graph. The entities under discussion are associated to nodes in the dialogue state graph. Orange represents \textcolor{Orange}{location} entities, green represents \textcolor{Green}{people} entities, and purple is \textcolor{Purple}{events}.}
    \label{fig:dialogue1}
\end{figure*}

\begin{figure*}[t]
    \FramedBox{5.5cm}{11.25cm}{
    \textbf{1} \textsc{Visitor:}\ \ Hello I am \textcolor{Green}{\dotuline{Andrew Fletcher}} \\
    \textbf{2} \textsc{Robot:}\ \ \ \ \ Hello \textcolor{Green}{\dotuline{Andrew}}! How can I help you? \\
    \textbf{3} \textsc{Visitor:}\ Do you know if \textcolor{Brown}{\dotuline{Stephanie Jules}} is attending a meeting called \textcolor{Purple}{\dotuline{infrastructures discussion}} \\
    \textbf{4} \textsc{Robot:}\ \ \ \ \ I can't find anyone named \textcolor{Brown}{\dotuline{Stephanie Jules}}, did you mean \textcolor{Green}{\dotuline{Stephanie Shields}}? \\
    \textbf{5} \textsc{Visitor:}\ \ [UNK] \\
    \textbf{6} \textsc{Robot:}\ \ \ \ \ Sorry, I didn't understand that. Could you repeat? \\ 
    \textbf{7} \textsc{Visitor:}\ \ Yes I meant \textcolor{Green}{\dotuline{Stephanie Shields}} \\
    \textbf{8} \textsc{Robot:}\ \ \ \ \ \textcolor{Green}{\dotuline{Stephanie Shields}} is not attending \textcolor{Purple}{\dotuline{infrastructures discussion}}. \\ 
    \textbf{9} \textsc{Visitor:}\ \ Could you add \textcolor{Green}{\dotuline{her}} to \textcolor{Purple}{\dotuline{infrastructures discussion}} \\
    \textbf{10} \textsc{Robot:}\ \ \ \ Ok! I've sent \textcolor{Green}{\dotuline{her}} an invitation to \textcolor{Purple}{\dotuline{the event}}. \\ 
    \textbf{11} \textsc{Visitor:}\ Thank you very much \\
    \textbf{12} \textsc{Robot:}\ \ \ \ You're welcome! Is there anything else I can help you with? \\ 
    \textbf{13} \textsc{Visitor:}\ \ No that will be all \\
    \textbf{14} \textsc{Robot:}\ \ \ \ Okay! Have a nice day! 
    }
    \caption{Dialogue example including a misunderstanding due to ASR errors. The graph is updated on line 10 to add the individual "Stephanie Shields" as a participant to the event (creating an edge between her and the event). Green entities are \textcolor{Green}{people}, Brown are \textcolor{Brown}{spurious entities} not found in the dialogue state, and purple are \textcolor{Purple}{events}.}
    \label{fig:dialogue2}
\end{figure*}

Each dialogue is paired with the associated fictive organization generated for that dialogue, along with the task presented to the human user. The fictive organizations contain on average 90 entities of the types \{\emph{people, event, room, group}\}, along with start and end times for events and additional attributes such as emails and phone numbers which can in principle also be treated as distinct entities.

The raw recordings of each user turn are provided as WAV files along with Automatic Speech Recognition and "gold standard" transcriptions of the utterance. Each user turn was annotated with an abstract representation of the user intent, along with entity mentions in both ASR and gold transcriptions. The wizard responses were similarly annotated with a logical representation of the system actions along with entity mentions. Both user intents and agent actions are expressed with a flat predicate taking entities as arguments. The annotation of entity mentions describe entities that exist in the graph and can be linked from the utterance as well as spurious entities which cannot. Along with these links, we also annotate links between mentions referring to the speaker or a new event with a special token indicating that the link refers to a newly created entity in the task. As part of the calendar scheduling task, nodes are added to represent a newly created event for the user. Mentions from both user and system utterances are provided with annotations of which text span is the entity mention and which entity is the correct target. In the case of ambiguous names (e.g. two events with the same name on different days), the target entity is annotated by constraints describing the attributes of the target entity.

The annotations in GraphWOZ are provided in JSON format. The data is split into approximately 50 \% for training, 25 \& for development, and 25 \% for the held-out test set.



\section{Approach}
\label{sec:approach}

\subsection{Graph representation}

\begin{figure*}[t]
\centering
    \includegraphics[width=.76\textwidth]{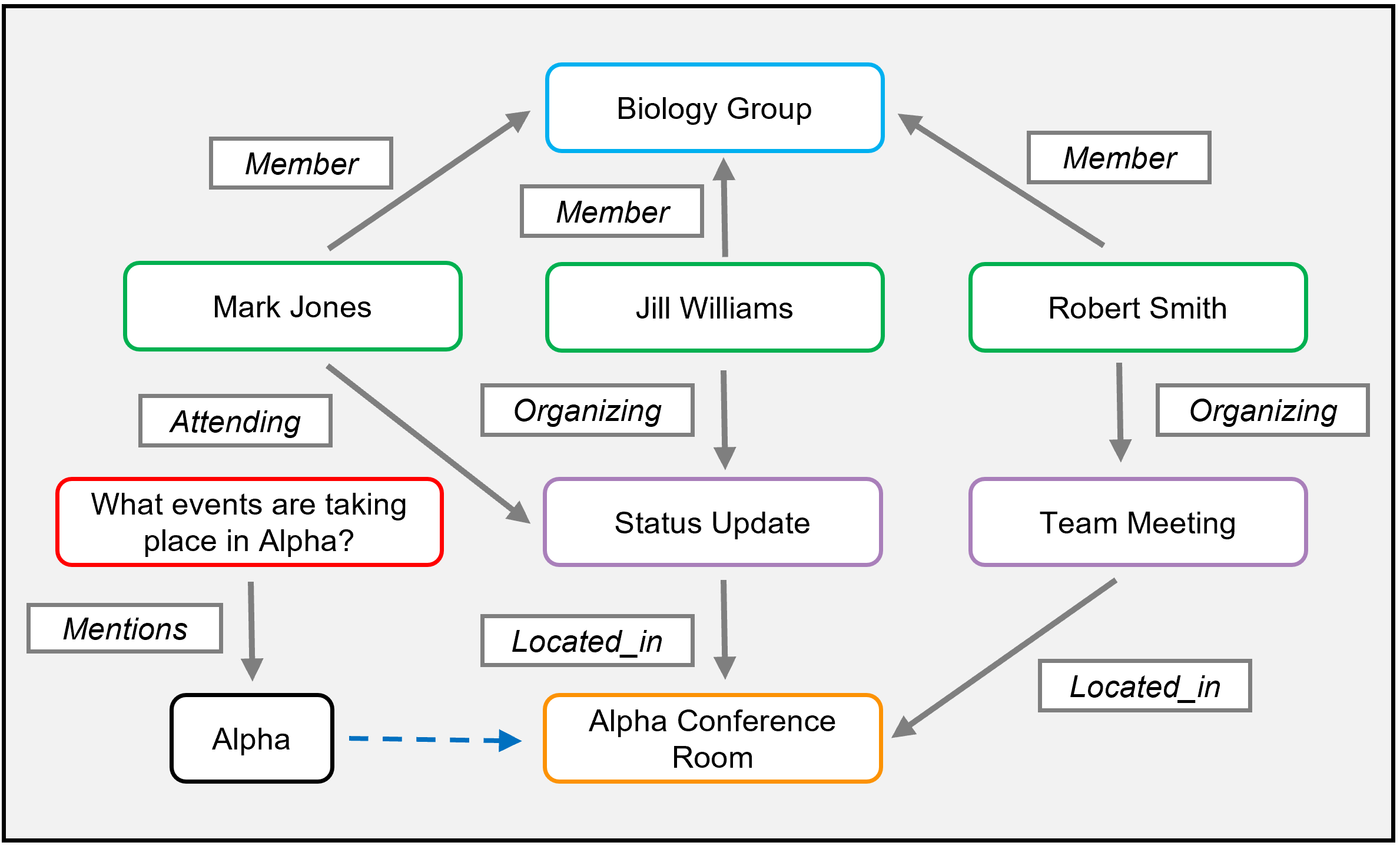}
    \caption{Example (directed) dialogue graph with entity linking. Nodes represent people, places, or dialogue elements, and edges represent semantic relations between them. Green nodes are \textcolor{Green}{people} entities, purple nodes are \textcolor{Purple}{events}, orange is a \textcolor{Orange}{location}, light blue is a \textcolor{cyan}{group}, and red is a \textcolor{Red}{user utterance}. Edges are labeled as semantic relations. The dark blue dashed arrow represents a \textcolor{Blue}{link created by entity linking}, recognizing that "Alpha" refers to the Alpha Conference Room.}
    \label{fig:graphexample}
\end{figure*}

Our approach to dialogue management centers around a dynamic entity knowledge graph. The graph represents both general background knowledge for the interaction (in this case events, persons, groups and contact information known to the robot receptionist) but also nodes and edges representing the dialogue itself (such as utterances, intents, responses, and entity mentions). The relations contained in the dataset are expressed through directed edges, and each node is associated with a semantic type.  A small example of state graph is provided in Figure \ref{fig:graphexample}.

In this setting, dialogue management centers on graph transformations which alter the graph of the dialogue state. Each new utterance is represented as a new node in the graph, with attributes including the ASR transcription itself as well as start and end timestamps. These utterances are added to the graph along with a linked node for each entity mention they contain, along with a link to the preceding utterance node (if any). The link between utterances represents a sequential relation between them. At this stage, the dialogue management system must determine which (if any) entities from the existing knowledge graph refer to the same entity as the entity mention. For instance, if a person is mentioned, they may be identified by name, a pronoun, or some other construction (e.g. "that person"). This identification step can readily be framed as an entity linking task. 

To construct the graph from the annotated data, we consider each entity of type \{\emph{people, event, room, group}\} as an individual node in the graph. For this purpose, a person's office is represented as a room as well, although it could equally be viewed as an attribute of the person entity. All nodes can in the simplest case, and for our experiments, be assigned the entity name as their attribute. Alternatively, these strings or a combination of attributes such as a person's email or phone number can in principle also be encoded with a language model to assign a vector representation to each node. Similarly, the edges of the graph are assigned labels to represent the semantic relationship between them. For example, each event in the data has a set of attendees and an organizer, thus each person in these attributes is assigned a link to the relevant link node. For our experiments, the task of entity linking can be viewed as predicting the occurrence of the \begin{small}\textsf{refers\_to}\end{small} relationship between an entity mention (text span present in an utterance) and the actual entity it refers to in the background knowledge of the robot receptionist. Other semantic relations such as  \begin{small}\textsf{attending}\end{small} or  \begin{small}\textsf{is\_located}\end{small} capture relationships that are valuable for spatio-temporal reasoning.

The knowledge graph also represents the set of user and agent utterances as nodes. Each node contains the raw string of the utterance (the ASR transcription for the user utterance) and is connected to nodes representing entity mentions in the utterance. An entity \textit{mention} is thereby treated separately from whichever known entities it corresponds to in the graph, where an unlinked entity mention may be a spurious entity that cannot be connected to the known set of entities. Additionally, an underspecified entity links to multiple entities in the graph, where the exact reference should be clarified by the system at a later round.

\subsection{Entity Linking}
\label{entity_linking}

Along with the release of the GraphWOZ dataset, we present simple but effective approaches for two specific tasks: entity-linking and response selection.

Upon recognizing a new user utterance, a graph-based dialogue system must link entities recognized in the user utterance with any matching entities in the graph. For each mention $m \in M$ and a set of known entities $E$, we wish to link $m$ to a set of entities $E' \subset E$.
For our experiments, we consider all dialogue turns containing at least one mention where $E' \neq \emptyset$. Note that, in contrast to traditional entity-linking models, we allow a mention to connect to more than one entity, as is the case for e.g. plural pronouns.

\subsubsection*{Input Features} 

The following features were employed for the entity-linking model: 

\textbf{String distance}: Many entity mentions in a dialogue refer to the entity directly by name. In these cases, it is appropriate to simply search for the the recognized entity string to the entities in the graph. For this purpose, we use string equality as a simple binary feature to the model. However, due to ASR error the transcription of the entity may be corrupted and require edit distance metrics to recognize the link. To this end, our baseline entity linking model employs several such metrics, namely: \emph{Levenshtein Distance, Jaro-Winkler Distance}, and \emph{Longest-Common Substring}. We additionally add features which are also calculated from the edit-distance features divided by the length of the mention string (to account for pronouns). Additionally, we compare similarity on the token level by computing the set difference between the tokens in the mention and candidate target. Similarly, we also we also add the cardinality of the union of tokens between the pair as a feature.

\textbf{Graph distance}: At a given turn in the dialogue, there may exist previous entity mention nodes which have been linked to another entity in the graph. Upon addition of a new mention node to the graph, it is necessary to also consider whether the new node refers to a previously mentioned entity.  The graph structure is useful to this purpose, as the distance to previously mentioned nodes corresponds directly to how recently they were mentioned. The graph distance can then be used as an indication of relevance and recency of possible targets for mentions at the current turn.  


A new entity mention may in many cases refer back to a node which is more distant than other intervening mention-entity pairs, so it is desirable to evaluate whether the two mentions are likely part of the same \emph{coreference chain}. In this way, entity linking in this graph can be viewed as a task of coreference resolution where we can also make use of linguistic structure to assist the model predictions. From this intuition, we use two pretrained coreference resolution models to extract coreference chains from the dialogue utterances:
\begin{enumerate}
    \item The first model is based on token-level coreference resolution as implemented in SpaCy Coreferee\footnote{\url{https://github.com/msg-systems/coreferee}}, and relies on contextualized word representations extracted via RoBERTa \cite{DBLP:journals/corr/abs-1907-11692}.
    \item The second model employs the end-to-end coreference resolution method described by \cite{lee-etal-2017-end}, as implemented in Allen NLP\footnote{\url{http://docs.allennlp.org/v0.9.0/api/allennlp.models.coreference\_resolution.html}}, and using SpanBERT embeddings \cite{joshi-etal-2020-spanbert} to derive the coreference clusters.  
\end{enumerate}

Based on the coreference chains extracted with those models, we add a link between all entity mentions which exist in the same coreference chain. This procedure greatly shortens the graph distance from a new mention node to the correct target when it refers to a previously mentioned entity.

\subsubsection*{Classification model} 

The string and graph-based distance features described above are then used as inputs to train binary classification models for the task of predicting whether a (mention, entity) pair should be linked together. Given the modest size of the training set, those classification models need to be limited to a small set of parameters. In Section \ref{sec:results}, we present results obtained with a simple logistic regression model and a feedforward neural model with two hidden layers. 

\subsection{Response Ranking}

Once entity mentions are linked to their corresponding entity in the graph-structured dialogue state, we proceed with the task of selecting the system response. The approach rests upon determining which part of the graph is relevant for selecting the next action and then converting this subgraph into a short text. This text is then prepended to the dialogue history and used to rank possible responses using a fine-tuned language model. 

The response ranking approach relies on the following steps:

\subsubsection*{Extraction of relevant subgraph} 
We first extract a relevant subgraph of the dialogue state to filter out entities and relations that are unlikely to influence the response selection. To this end, we first determine the entities explicitly mentioned through the dialogue history, along with the nodes adjacent to those mentioned entities, and the set of all relations between those nodes. These entities are determined using the output of the MLP-based entity linking system described in the previous section.


\subsubsection*{Conversion of subgraph into NL sentences}
\begin{figure}
    \includegraphics[width=1\textwidth]{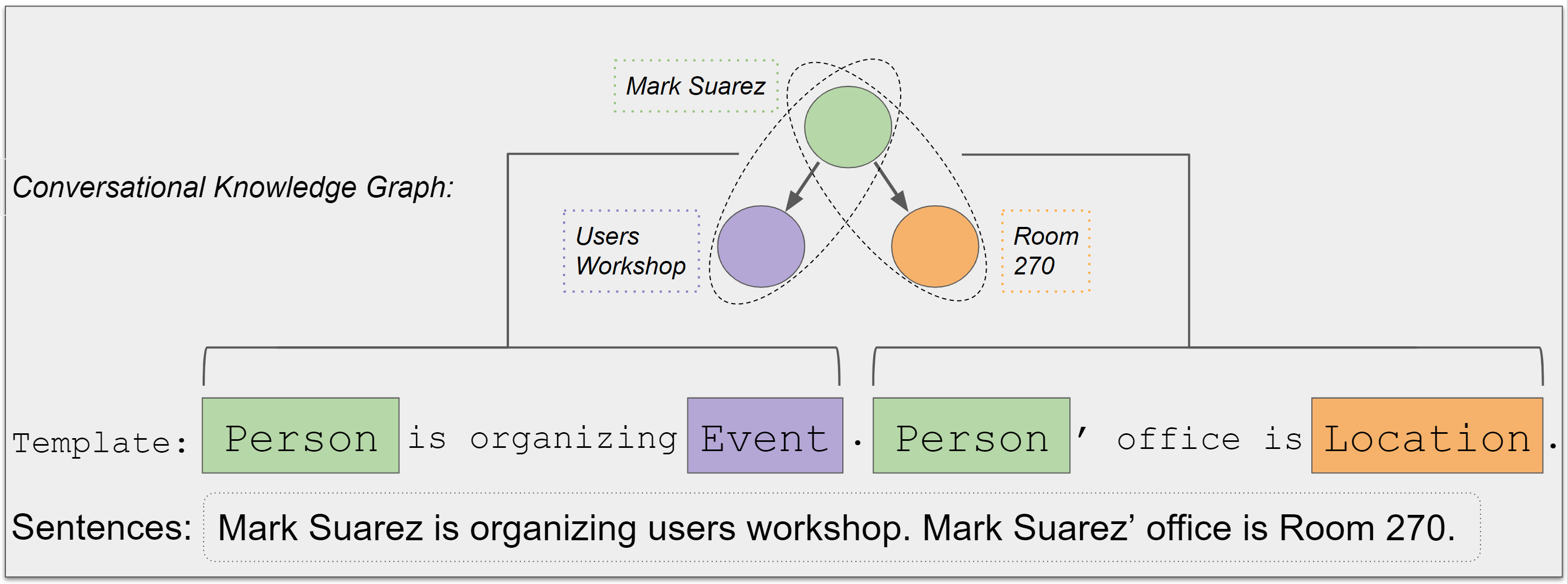}
    \caption{Example of two sentences filled in by a template. These sentences along with others filled by templates describing entity attributes are combined into a document describing a person. The document created with these templates is combined with the dialogue history to form the context for the language model.}
    \label{fig:template}
\end{figure}

In order to take advantage of large pretrained language models, the subgraph is then converted from a graph representation into text using predefined templates. Once generated, the text is then prepended to the dialogue history. As an example, the relevant subgraph containing entities mentioned in the dialogue in Figure \ref{fig:dialogue1} are converted into two sentences, as illustrated in Figure \ref{fig:template}. 


\subsubsection*{Fine-tuning of ranking model}

Finally, we fine-tune a neural language model. Two alternative approaches are evaluated: \begin{itemize}
    \item A pointwise approach taking a (context, response) pair as input and returning a normalized score denoting the appropriateness of the response given the conversational context.  
    \item A pairwise approach where the model is simultaneously provided with a positive pair (context, actual response) and a negative pair (context, random response). The model is then optimized to minimize the cross-entropy loss for the difference between the positive and negative examples \cite{burges2005learning}. 
\end{itemize}

To mitigate the scarcity of training data, we expand the initial set of (dialogue context, wizard response) pairs using a simple mechanism to swap entity names for persons, events and groups with alternative values from the knowledge graph. Those modified pairs are then added to the training set. 

\textit{Implementation details:} The neural ranking model is fine-tuned using a BERT language model \cite{devlin2018bert} on 10 epochs (batch size of 5) with a cross-entropy loss and the Adam optimizer \cite{kingma2014adam} with a learning rate set to $1 \times 10^{-6}$. For the negative examples, we experimented both with a random selection of responses and with a more informed approach where candidates are samples according to their cosine similarity, using dense embeddings from \cite{reimers-2019-sentence-bert}. Those two negative sampling methods led to similar experimental results on GraphWOZ.

\section{Results}
\label{sec:results}

\subsection{Entity-linking}

Two entity-linking models were tested, respectively based on logistic regression and a multilayer perceptron (MLP). For each (mention, entity) pair, we extract both string and graph features as described in section \ref{entity_linking} and provide them as input to the classifier. For the MLP model, the best performing parameters were two layers of size 20 and 12, trained for 600 iterations optimized with Limited-memory BFGS. Those hyper-parameters were selected empirically based on the development set. For comparison, we also provide two simple baselines:
\begin{itemize}
    \item \textit{String equality}, which connects an entity mention to an entity represented by the exact same string.
    \item \textit{String equality + Recency}, which first checks the existence of an entity of same name as above. In case no exact string match can be found, the baseline then connects the entity mention to the most recently mentioned entity. 
    \end{itemize}

Table \ref{tab:entity_results} shows the evaluation results of the entity linking models with precision, recall, and $F_1$-score. We observe that both the logistic regression and feedforward neural model benefit from the inclusion of graph-distance features. It can also be noted that the string equality baseline achieves less than perfect precision due to ambiguity between different entities with the same name. 


\begin{table}
\centering
  \begin{tabular}{lSSSSSS}
    \toprule
    \multirow{2}{*}{Model} &
      \multicolumn{3}{c}{\textbf{Dev}} &
      \multicolumn{3}{c}{\textbf{Test}} \\
      & {Precision} & {Recall} & {F1} & {Precision} & {Recall} & {F1}\\
      \midrule
    String Equality & 0.92 & 0.22 & 0.35 & 0.96 & 0.32 & 0.48 \\ 
    String equality + Recency & 0.60 & 0.28 & 0.38 & 0.59 & 0.38 & 0.46 \\
    Logistic Regression, String & 0.85 & 0.33 & 0.48 & 0.89 & 0.46 & 0.61 \\
    Logistic Regression, String + Graph & 0.87 & 0.46 & 0.61 & 0.95 & 0.59 & 0.73\\
    MLP (best), String & 0.87 & 0.53 & 0.65 & 0.89 & 0.59 & 0.71 \\
    MLP (best), String + Graph & 0.89 & 0.64 & 0.74 & 0.88 & 0.67 & 0.76\\
    \bottomrule
  \end{tabular}
\caption{Entity Linking with string equality and recency heuristics, Logistic Regression, and MLP. Graph distance is computed as $\frac{1}{2^l}$ where $l$ is the length of the path from the mention to the target node. If there is no path, the $\frac{1}{2^{10}}$ is used (an intermediate but finite value $l$, as many correct utterance links will have no existing path between the nodes).}
\label{tab:entity_results}
\end{table}



\subsection{Response ranking}

Table \ref{tab:my_label} provides the evaluation results on the task of ranking possible system responses using standard IR metrics, namely the recall $R_{10}@1$, which measures how often the correct response is ranked highest among a set of $10$ candidate responses, the recall $R_{10}@2$, which measures how often the correct response is placed in the top-2 among 10 candidates, and the Mean Reciprocal Rank (MRR), which is the harmonic mean of the rank of the correct response. Those results are obtained using the best entity-linking model from the previous section (MLP with both string and graph distance features) to link entity mentions to their corresponding entities.

As we can observe from Table \ref{tab:my_label}, the performance of the response ranking models (both pointwise and pairwise) improves when the content of the relevant subgraph (converted to a text format through templates) is prepended to the dialogue history. The pairwise model also seems to perform slightly better than the pointwise variant. 

\begin{table*}[h]
    \centering
    \begin{tabular}{p{35mm}p{22mm}p{8.5mm}p{8.5mm}p{8.5mm}p{8.5mm}p{8.5mm}p{8.5mm}}
    \toprule
    Input & Ranking model & \multicolumn{2}{c}{$R_{10}@1$} & \multicolumn{2}{c}{$R_{10}@2$} & \multicolumn{2}{c}{MRR} \\
    \cmidrule(l{1pt}r{1pt}){3-4}\cmidrule(l{1pt}r{1pt}){5-6}\cmidrule(l{1pt}r{1pt}){7-8}
    & & Dev & Test &Dev & Test &Dev & Test \\ \midrule
    \multirow{2}{5cm}{Only dialogue history} & Pointwise & 0.76 & 0.72 & 0.91 & 0.91 & 0.85 & 0.83 \\
      &  Pairwise & 0.80 & 0.83 & 0.89 & 0.94 & 0.87 & 0.90 \\ \cmidrule{1-1}
    \multirow{2}{35mm}{Relevant subgraph (in text form) + dialogue history} & Pointwise & 0.80 & 0.85 & 0.91 & 0.93 & 0.88 & 0.91 \\
    & Pairwise  & 0.85 & 0.84 & 0.91 & 0.95 & 0.95 & 0.91  \\

    \bottomrule
    \end{tabular}
    \caption{Evaluation results on the development and test sets of the GraphWOZ dataset for the task of selecting the next agent response. The results are computed based from a 10 candidate responses for each dialogue turn. In addition to the type of ranking model (pointwise or pairwise), we also show results either with or without the natural language sentences summarizing the content of the relevant subgraph for the dialogue history.}
    \label{tab:my_label}
\end{table*}

\section{Conclusion}
\label{sec:conclusion}


This paper presented GraphWOZ, a new collection of Wizard-of-Oz dialogues where the dialogue state is structured as a graph including a varying number of entities (persons, events, groups, etc.) and relations between them. The dialogues were gathered as part of a human--robot interaction scenario where the robot was expected to operate as a receptionist and answer various requests regarding the agenda of a fictive organization. Each dialogue is coupled with a randomly generated knowledge graph, and is provided with a range of annotations, notably the intents (expressed as logical forms) associated with each user utterance and system response as well as the entities mentioned through the dialogue. The GraphWOZ dataset comprises a small but richly annotated set of dialogues in a novel domain which explicitly represents the dialogue state as a single, unified graph.

In addition to this dataset, we also present simple but effective approaches to two dialogue management tasks (conversational entity linking and response ranking) based on a graph-structured dialogue state.  For entity-linking, experimental results on GraphWOZ demonstrate that the use of both string and graph distance features allows the classification model to perform entity linking for both direct references as well as pronouns and other indirect references.

We also provide empirical results on the task of response selection based on the dialogue state graph and dialogue history. The presented approach uses the results of the entity linking model to select a relevant subgraph based on the entities linked from user utterances. This subgraph is then converted into natural language sentences and prepended to the dialogue context. The resulting text is finally fed to a neural language model and employed to rank candidate responses.

The response selection model presented in this paper depends on the availability of templates that can express the content of the dialogue state graph into English sentences. Although we found those templates to be relatively easy to create for this particular domain, this requirement may be difficult to satisfy for other types of graph. Future work will focus on investigating the use of graph neural networks to perform dialogue state tracking and response selection on the GraphWOZ dataset without the need to convert the graph into text. Additionally, the ability of graphs to represent uncertainty by means of e.g. weighting of proposed entity links is a potential topic for exploration.

\newpage


\bibliographystyle{styles/spmpsci}
\bibliography{biblio.bib}


%

\backmatter
\appendix
%
%
%

\chapter{Chapter Heading}
\label{introA} 

Use the template \emph{appendix.tex} together with the Springer document class SVMono (monograph-type books) or SVMult (edited books) to style appendix of your book in the Springer layout.

\section{Section Heading}
\label{sec:A1}
Instead of simply listing headings of different levels we recommend to let every heading be followed by at least a short passage of text. Further on please use the \LaTeX\ automatism for all your cross-references and citations.

\subsection{Subsection Heading}
\label{sec:A2}
Instead of simply listing headings of different levels we recommend to let every heading be followed by at least a short passage of text. Further on please use the \LaTeX\ automatism for all your cross-references and citations as has already been described in Sect.~\ref{sec:A1}.

For multiline equations we recommend to use the \verb|eqnarray| environment.
\begin{eqnarray}
\vec{a}\times\vec{b}=\vec{c} \nonumber\\
\vec{a}\times\vec{b}=\vec{c}
\label{eq:A01}
\end{eqnarray}

\subsubsection{Subsubsection Heading}
Instead of simply listing headings of different levels we recommend to let every heading be followed by at least a short passage of text. Further on please use the \LaTeX\ automatism for all your cross-references and citations as has already been described in Sect.~\ref{sec:A2}.

Please note that the first line of text that follows a heading is not indented, whereas the first lines of all subsequent paragraphs are.

%
\begin{figure}[t]
\sidecaption[t]
\includegraphics[scale=.65]{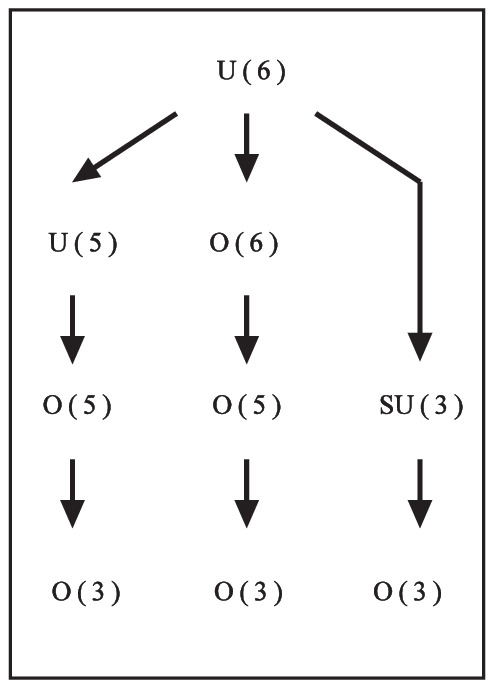}
%
%
\caption{Please write your figure caption here}
\label{fig:A1}       
\end{figure}

%
\begin{table}
\caption{Please write your table caption here}
\label{tab:A1}       
%
%
\begin{tabular}{p{2cm}p{2.4cm}p{2cm}p{4.9cm}}
\hline\noalign{\smallskip}
Classes & Subclass & Length & Action Mechanism  \\
\noalign{\smallskip}\hline\noalign{\smallskip}
Translation & mRNA$^a$  & 22 (19--25) & Translation repression, mRNA cleavage\\
Translation & mRNA cleavage & 21 & mRNA cleavage\\
Translation & mRNA  & 21--22 & mRNA cleavage\\
Translation & mRNA  & 24--26 & Histone and DNA Modification\\
\noalign{\smallskip}\hline\noalign{\smallskip}
\end{tabular}
$^a$ Table foot note (with superscript)
\end{table}
%

%
%

\Extrachap{Glossary}

Use the template \emph{glossary.tex} together with the Springer document class SVMono (monograph-type books) or SVMult (edited books) to style your glossary\index{glossary} in the Springer layout.

\runinhead{glossary term} Write here the description of the glossary term. Write here the description of the glossary term. Write here the description of the glossary term.

\runinhead{glossary term} Write here the description of the glossary term. Write here the description of the glossary term. Write here the description of the glossary term.

\runinhead{glossary term} Write here the description of the glossary term. Write here the description of the glossary term. Write here the description of the glossary term.

\runinhead{glossary term} Write here the description of the glossary term. Write here the description of the glossary term. Write here the description of the glossary term.

\runinhead{glossary term} Write here the description of the glossary term. Write here the description of the glossary term. Write here the description of the glossary term.
\printindex


\end{document}